Article

# An Optimal House Price Prediction Algorithm: XGBoost

Hemlata Sharma [1], Hitesh Harsora [1] and Bayode Ogunleye [2,*]

1. Department of Computing, Sheffield Hallam University, Sheffield S1 2NU, UK; h.sharma@shu.ac.uk (H.S.)
2. Department of Computing & Mathematics, University of Brighton, Brighton BN2 4GJ, UK
* Correspondence: b.ogunleye@brighton.ac.uk

**Abstract:** An accurate prediction of house prices is a fundamental requirement for various sectors, including real estate and mortgage lending. It is widely recognized that a property's value is not solely determined by its physical attributes but is significantly influenced by its surrounding neighborhood. Meeting the diverse housing needs of individuals while balancing budget constraints is a primary concern for real estate developers. To this end, we addressed the house price prediction problem as a regression task and thus employed various machine learning (ML) techniques capable of expressing the significance of independent variables. We made use of the housing dataset of Ames City in Iowa, USA to compare XGBoost, support vector regressor, random forest regressor, multilayer perceptron, and multiple linear regression algorithms for house price prediction. Afterwards, we identified the key factors that influence housing costs. Our results show that XGBoost is the best performing model for house price prediction. Our findings present valuable insights and tools for stakeholders, facilitating more accurate property price estimates and, in turn, enabling more informed decision making to meet the housing needs of diverse populations while considering budget constraints.

## 1. Introduction

Housing is one of the basic human needs. House price prediction is of utmost importance for real estate and mortgage lending organizations due to the significant contribution of the real estate sector to the global economy. This process is beneficial not only for businesses but also for buyers, as it helps mitigate risks and bridges the gap between supply and demand [1]. To estimate house prices, regression methods are commonly employed, utilizing numerous variables to create models [2]. An efficient and accessible housing price prediction model has numerous benefits for various stakeholders. Real estate businesses can utilize the model to assess risks and make informed investment decisions. Mortgage lending organizations can leverage it to evaluate loan applications and determine appropriate interest rates. Buyers can use the model to estimate the affordability of properties and make informed purchasing decisions. Most importantly, the recent instability of house prices has made the need for prediction models more important than before.

Previous studies [3–5] have applied various machine learning (ML) algorithms for house price prediction, with the focus on developing a model; not much attention has been paid to house price predictors. The literature synthesis discussed in Section 2 shows that only a limited number of studies have explicitly discussed the influential factors impacting model performance. Our literature review findings suggest that various traditional ML algorithms have been studied; however, there is a need to identify the optimal methodology for house price prediction. For example, Madhuri et al. [6] compared multiple linear regression, lasso regression, ridge regression, elastic net regression, and gradient boosting regression algorithms for house price prediction. However, their study did not propose an optimal solution. This is due to the fact that they applied regression algorithms using the default settings only, with no attempt to achieve optimality. The dearth of research regarding this underscores the need for a more comprehensive study on the diverse elements that contribute to the effectiveness of house price predictive models. By delving deeper into the identification and analysis of these influential factors, we can unveil valuable insights that will aid in achieving an optimal house price prediction model. This is beneficial to the real estate sector for understanding the significant factors that influence house costs.

Thus, this study aims to develop a house price prediction model and also identify the significant factors that influence house price prediction. To this end, this study formulates this problem as a regression task and thus conducts a comprehensive experimental comparison of ML techniques to ascertain the most effective model that accurately predicts house prices. The findings of this study not only provide insights into the comparative effectiveness of different ML techniques but also contribute valuable information for selecting optimal models based on specific data characteristics. Specifically, the experimental results propose the use of the XGBoost algorithm based on its interpretability, simplicity, and performance accuracy. XGBoost is a widely used ML algorithm, and the impact of the algorithm has been praised in a number of machine learning and data mining problems based on scalability, interpretability, and applicability [7].

The rest of this paper is structured into four main parts. Section 2 centers on the literature review, Section 3 highlights the method and evaluation process, the next section presents the result, while the last section concludes and suggests future avenues to consider.

## 2. Related Work

Predicting house prices provides insights into economic trends, guides investment decisions, and supports the development of effective policies for sustainable housing markets. The study by [8] emphasized the reliance of real estate investors and portfolio managers on house price predictions for making informed investment decisions. Recent market trends have demonstrated a clear connection between the accuracy of these predictions and the improved optimization of investment portfolios. Anticipating fluctuations in house prices empowers investors to proactively adapt their portfolios, seize emerging opportunities, and strategically navigate risks, leading to more robust and resilient investment outcomes. Furthermore, the authors in [9] discussed how individuals can gain a better understanding of real estate for their own personal investment and financing decisions. Similarly, Ref. [10] added that financial institutions and policymakers recognize house price trends as an economic indicator, as it is important to note that fluctuations in house prices can affect consumer spending, borrowing, and the overall economy. The study by [11] proposed an intuitive theoretical model of house prices, where the demand for housing was driven by how much individuals could borrow from financial institutions. A borrower's level of debt depends on the level of disposable income he or she has and the current interest rate. The study showed that actual house prices and the amount individuals can borrow are related in the long run with plausible and statistically significant adjustments. The authors in [12] argued that landscape influences the real estate market, adding that macro- (foreign exchange) and micro-variables (such as transportation access, financial stability, and stocks) can change the land price, therefore these can be used to predict future land prices.

ML has revolutionized the process of uncovering patterns and making reliable predictions. This is due to the fact that ML involves the process of acquiring knowledge from past experiences in relation to specific tasks and performance criteria [13]. ML algorithms are of two main categories, namely the supervised and the unsupervised ML approach [14]. The supervised ML approach makes use of a subset of labeled data (where target variable is known) for training and testing on the remaining data to make predictions on unseen datasets [15]. Whilst the unsupervised ML approach does not require a labeled dataset, the approach facilitates the analysis (by uncovering hidden patterns) and makes prediction from unlabeled datasets [16]. In the context of house price prediction, previous studies have conceptualized the problem as a classification task [17] or a regression task [18]. The supervised ML algorithms are capable of modeling both tasks. An example of the classi-fication approach was performed in the work of [17]. They aimed to predict whether the closing house price was greater than or less than the listing house price. They transformed the target variable as "*high*" when the closing price was greater than or equal to the listing price and as "*low*" when the closing price was lower than the listing price. Thus, their classification result showed that RIPPER (repeated incremental pruning to produce error reduction) outperformed C4.5, naïve Bayes, and AdaBoost in the Fairfax County, Virginia house dataset, which consisted of 5359 townhouse records.

Most studies have approached the house price prediction problem as a regression task to be able to provide estimates that are predictive in determining the direction of future trends. For example, in China, Ref. [19] used 9875 records of Jinan city estate market data for house price prediction. The paper showed that CatBoost was superior to multiple linear regression and random forest, with an R-squared of 91.3% and an RMSE of 772.408. In the Norwegian housing market, Ref. [20] introduced squared percentage error (SPE) loss function to improve XGBoost for a house price prediction model. Thus, they showed that their SPE loss function XGBoost algorithm—named SPE-XGBoost—achieved the lowest RMSE of 0.154. The authors in [18] used a Boston (USA) house dataset that consisted of 506 entries and 14 features to implement a random forest regressor and achieved an R-squared of 90%, an MSE (mean square error) of 6.7026, and an RMSE (root mean square error) of 2.5889. Similarly, Ref. [21] showed that lasso regression outperformed linear regression, polynomial regression, and ridge regression using the Boston house dataset, with an R-squared of 88.79% and an RMSE of 2.833. The authors in [6] used the King County housing dataset to compare multiple linear regression, ridge regression, lasso regression, elastic net regression, AdaBoost regression, and gradient boosting, and showed that gradient boosting achieved the superior result. However, it is worth stating that most of these studies applied a basic (default) regression model without considering optimizing the model and did not perform a comprehensive analysis of the feature importance. For illustration, this study provides a summary of the literature findings in Table 1 below.

**Table 1.** Summary of the literature evidencing dataset used and their findings.

| Author | Dataset | Findings | RMSE |
|---|---|---|---|
| Zou [19] | Jinan city estate market, China | CatBoost is superior to multiple linear regression and random forest, with an R-squared of 91.3%. | 772.408 |
| Hjort et al. [20] | Norwegian housing market | SPE-XGBoost achieved the lowest RMSE compared with linear regression, nearest neighbour regression, random forest, and SE-XGBoost. | 0.154 |
| Adetunji et al. [18] | Boston (USA) house dataset | Random forest regressor achieved an R-squared of 90% and an MSE (mean square error) of 6.7026. | 2.5889 |
| Sanyal et al. [21] | Boston (USA) house dataset | Lasso regression outperformed linear regression, polynomial regression, and ridge regression with an R-squared of 88.79%. | 2.833 |
| Madhuri et al. [6] | King County housing (USA) | Gradient boosting showed a superior result with an adjusted R-squared of 91.77% over multiple linear regression, ridge regression, lasso regression, elastic net regression, and AdaBoost regression. | 10,971,390,390 |
| Aijohani [1] | King County housing (USA) | Ridge regression outperformed lasso regression and multiple linear regression with an adjusted R-squared of 67.3%. | 224,121 |

**Table 1.** *Cont.*

| Author | Dataset | Findings | RMSE |
|---|---|---|---|
| Viana and Barbosa [22] | 1. King County (KC), USA; 2. Fayette Count (FC), USA; 3. São Paulo (SP), Brazil; 4. Porto Alegre (POA), Brazil. | Spatial interpolation attention network and linear regression showed robust performance over other models such as random forest, Lightgbm, XGboost, and auto-sklearn. | 115,763 (KC) 22,783 (FC) 154,964 (SP) 94,201 (POA) |

In summary, we reviewed the recent literature, specifically in the context of techniques utilized to provide up-to-date information on the house price prediction models. Our findings showed that only a few studies considered optimality and the significance of features. To evidence this, we summarized the techniques (including the optimization approach) that have been used in previous studies, as shown in Table 2 below.

**Table 2.** Summary of the recent literature evidencing techniques/optimization.

| Author(s) | Method | Hyperparameter Tuning |
|---|---|---|
| Azimlu et al. [23] | ANN, GP, Lasso, Ridge, Linear, Polynomial, SVR | Not performed |
| Wang [24] | OLS Linear Regression, Random Forest | Not performed |
| Fan et al. [25] | Ridge Linear Regression, Lasso Linear Regression, Random Forest, Support Vector Regressor (Linear Kernel and Gaussian Kernel), XGBoost | GridSearchCV |
| Viana and Barbosa [22] | Linear Regression, Random Forest, LightGBM, XGBoost, Auto-klearn, Regression Layer | Keras(Hyperas) |
| Aijohani [1] | Multiple Regression, Lasso Regression, Ridge Regression | Not performed |
| Sharma et al. [26] | Linear Regression, Gradient Boosting Regressor, Histogram Gradient Boosting Regressor, and Random Forest | Not performed |
| Madhuri et al. [6] | Multiple Regression, Lasso Regression, Ridge Regression, Elastic Net Regression, and Gradient Boosting Regression | Not performed |

## 3. Methodology

This section presents the methods employed for the house price prediction. In this paper, we compared several regression models, including linear regression (LR), multi-layer perceptron (MLP), random forest regression (RF), support vector regressor (SVR), and extreme gradient boosting (XGBoost) to ascertain the interpretable best performing model. We used the housing data from the Kaggle repository, which is publicly accessible via the link https://www.kaggle.com/datasets/shashanknecrothapa/ames-housing-dataset (accessed on 2 November 2023). The dataset consists of 2930 records (houses) with 82 variables. Furthermore, we discussed the regression techniques used to forecast house prices in the subsections below. This encompassed the selection of suitable ML algorithms evidenced in the literature.

### 3.1. Linear Regression

A simple and popular approach to house price prediction is linear regression (LR). LR is a statistical tool that establishes a relationship between a dependent variable (Y) and one or more independent variables (Xi). This relationship is represented by an equation in the form of

$$Y = \beta_0 + \sum_i^k \beta_i X_i + \epsilon \tag{1}$$

where $\beta_0$ is the intercept, $\beta_i$ are the slopes, $X_i$ are the independent variables, and $\epsilon$ is the error term.

*3.2. Random Forest*

Random forest (RF)—introduced by [27]—is a robust and versatile ensemble learning technique that offers precise results for various types of datasets [14,15,28,29]. Widely recognized for its precision in predicting outcomes for various datasets, RF excels in handling high-dimensional data, capturing complex relationships, and mitigating overfitting. Its effectiveness has been demonstrated across diverse domains such as financial forecasting and healthcare analytics. RF's innate resilience to overfitting and its ability to handle complex datasets make it a compelling choice in ML, offering precise and reliable predictions for a wide range of scenarios. It operates in a unique manner by giving less weight to weak features, resulting in faster processing compared with other methods. This characteristic makes it a reliable choice for handling missing or noisy data and outliers [14,15]. RF is versatile as it can tackle both classification and regression tasks. Additionally, it is capable of processing both categorical and continuous data types. A significant advantage of this model is that it is easy to interpret. RF overcomes the problem of overfitting by training multiple decision trees on different bootstrap samples from the training data [29]. The authors in [18] showed that RF is a powerful ML algorithm for house price prediction. However, a notable drawback of random forest is that it can become slow and inefficient when dealing with a large number of trees, making it less suitable for real-time predictions. Although the algorithm usually trains quickly, making predictions after training may take longer [30]. The RF predictor consists of an M randomized regression tree. Considering the $j^{th}$ tree in a cluster of trees, the predicted value at every query point x is denoted by $m_n(x; \varnothing_j, \partial_n)$, where $\varnothing_1, \ldots, \varnothing_m$ are the independent random variables and $\partial_n$ is the training variable [15]. The $j^{th}$ tree estimate is thus formulated as:

$$m_n(x; \varnothing_j, \partial_n) = \sum_{i \in \partial_n(\varnothing_j)} \frac{1_{X_i \in A_n(x;\varnothing_j,\partial_n)^{Y_1}}}{N_n(x; \varnothing_j, \partial_n)} \quad (2)$$

where $\partial_n^*(\varnothing_j)$ is the set of selected data points before tree construction.

$A_n(x; \varnothing_j, \partial_n)^{Y_1}$ is the cell containing x and $N_n(x; \varnothing_j, \partial_n)$ is the number of points selected before tree construction that fall into $A_n(x; \varnothing_j, \partial_n)^{Y_1}$. The finite forest estimate as a result of the combination of trees is then represented as:

$$m_{M,n}(x; \varnothing_1, \ldots, \varnothing_m, \partial_n) = \frac{1}{M} \sum_{j=1}^{m} m_n(x; \varnothing_j, \partial_n) \quad (3)$$

where M can take any size but is limited to computing resources.

*3.3. Support Vector Machines*

The support vector machine (SVM) is widely acknowledged and revered in the field of data mining and ML for its ability to effectively handle complex data patterns and achieve high-dimensional classification tasks with remarkable accuracy. In the 1990s, Vapnik [31] proposed SVM, which has since proven effective in ML applications. SVMs are versatile and capable of performing well in both classification and regression tasks.

In a classification task, SVM builds a hyperplane between separated marginal lines of the nearest support vectors (input vector). Within the space, an optimal separating hyperplane is determined by maximizing the marginal distance. The maximum margin hyperplane is used for optimal prediction. The higher the marginal distance, the more generalizable the result is. SVM is a linear classifier that also performs non-linear classification problems using the kernel function [14]. SVM offers precise prediction and is less susceptible to overfitting thanks to the utilization of the kernel trick [14,32]. The kernel trick is employed by SVMs to transform the data and determine the optimal decision boundary between potential outputs.

Similarly, for a regression task, the support vector regressor (SVR) is a class of SVM for regression. SVR aims to provide an estimate of a function $f(x)$. To this end, SVR fits the regression line in the E-insensitive region. It introduces an E-insensitive loss function to improve its out-of-sample performance. To obtain the regression function $f(x)$, SVR attempts to resolve the optimization problem illustrated below (where $e_i^*$ denotes slack variables (which is the error at the inbound of the E-insensitive region) and $e_i$ denotes slack variables (which is the error at the outbound of the E-insensitive region)).

Minimize:

$$\frac{1}{2}||kw^2|| + C \sum_{i=1}^{n} (e_i + e_i^*) \quad (4)$$

Subject to:

$$y_i - b - w'x_i \leq e + e_i$$
$$b + w'x_i - y_i \leq e + e_i^*$$
$$e_i \geq 0, \quad e_i^* \geq 0$$

### 3.4. Multi-Layer Perceptron (MLP)

Multi-layer perceptron (MLP) plays a crucial role as a fundamental building block in artificial neural networks (ANNs), offering a powerful tool for solving complex problems in various domains [33]. MLP, self-organizing map (SOM), and deep belief network (DBN) algorithm models are highly applicable when human experts are unavailable or unable to explain the decisions made using their knowledge adequately, when problem solutions evolve in size, and in instances where solutions must be modified depending on new information [34]. The authors in [35] highlighted the properties of MLP as capable of learning both linear as well as non-linear functions. MLP can learn how to perform tasks from the data given for training and initial experience. MLP minimizes the loss function. MLP is a stochastic program. Mathematically, the layers in a fully connected network, are formulated as shown in Equations (5)–(7) where the input units is denoted as $x_j$, $\varnothing^{(1)}$ and $\varnothing^{(2)}$ are the activation functions, $\omega$ represents the weights, units in the $l^{th}$ hidden layer are denoted as $h_i^{(l)}$ and the output y

$$h_i^{(1)} = \varnothing^{(1)} \left( \sum_j \omega_{ij}^{(1)} x_j + b_i^{(1)} \right) \quad (5)$$

$$h_i^{(2)} = \varnothing^{(2)} \left( \sum_j \omega_{ij}^{(2)} h_j^{(1)} + b_i^{(2)} \right) \quad (6)$$

$$y_i = \varnothing^{(3)} \left( \sum_j \omega_{ij}^{(3)} h_j^{(2)} + b_i^{(3)} \right) \quad (7)$$

### 3.5. Extreme Gradient Boosting

Extreme gradient boosting (XGBoost) has gained immense importance in the field of ML due to its exceptional performance and versatility. The authors in [7] introduced XGBoost as a scalable and efficient implementation of gradient tree boosting, where XGBoost could be used to solve real-world scale problems with a minimum number of resources. The open-source nature of XGBoost has fostered a collaborative community, leading to continuous improvements and innovations. Its popularity is also attributed to its resistance to overfitting, which is crucial for maintaining model generalization on new data [36]. The study by [37] applied XGBoost and multi-layer perceptron models to find out which one of them provides better accuracy and relationships between real-time variables. The study mentioned that XGBoost was more efficient than multi-layer perceptron for taxi trip duration-based predictions. The algorithm's success in diverse domains, including finance [38], healthcare [39], and supply chains [40], shows how popular the algorithm is. XGBoost can be mathematically represented as an optimization problem, where the objective function is formulated as below.

$$obj = \sum_{i=1}^{n} L(y_i, \hat{y}_i) + \sum_{k=1}^{K} \Omega(f_k) \quad (8)$$

where *K* is the number of trees, *L* ($y_i$, $\hat{y}_i$) is the loss function, and (*f*) is the regularization term that is used to control the complexity of the model and prevent the model from overfitting.

$$\Omega(f) = \gamma T + \frac{1}{2}\lambda \sum_{j=1}^{T} \omega_j^2 \qquad (9)$$

where *T* is the number of leaves and $\omega$ is the score of the leaf node. The predicted value of the $i^{th}$ sample after $t^{th}$ iteration is:

$$\hat{y}_i^{(t)} = \hat{y}_i^{(t-1)} + f_t(X_i) \qquad (10)$$

Also,

$$\sum_{i=1}^{t} \Omega(f_i) = \sum_{i=1}^{t-1} \Omega(f_i) + \Omega(f_i) \qquad (11)$$

$\sum_{i=1}^{t} \Omega(f_i)$ is a constant. From Equations (9)–(11):

$$obj^{(t)} = \sum_{j=1}^{T} \sum_{i \in I_j} L(y_i, \hat{y}_i^{(t-1)} + \omega_j) + \frac{1}{2}\lambda \omega_j^2 + \gamma T + constant \qquad (12)$$

### 3.6. Implementation

The implementation of an ML strategy involves multiple stages. Figure 1 below illustrates the key stages involved in conducting the ML process.

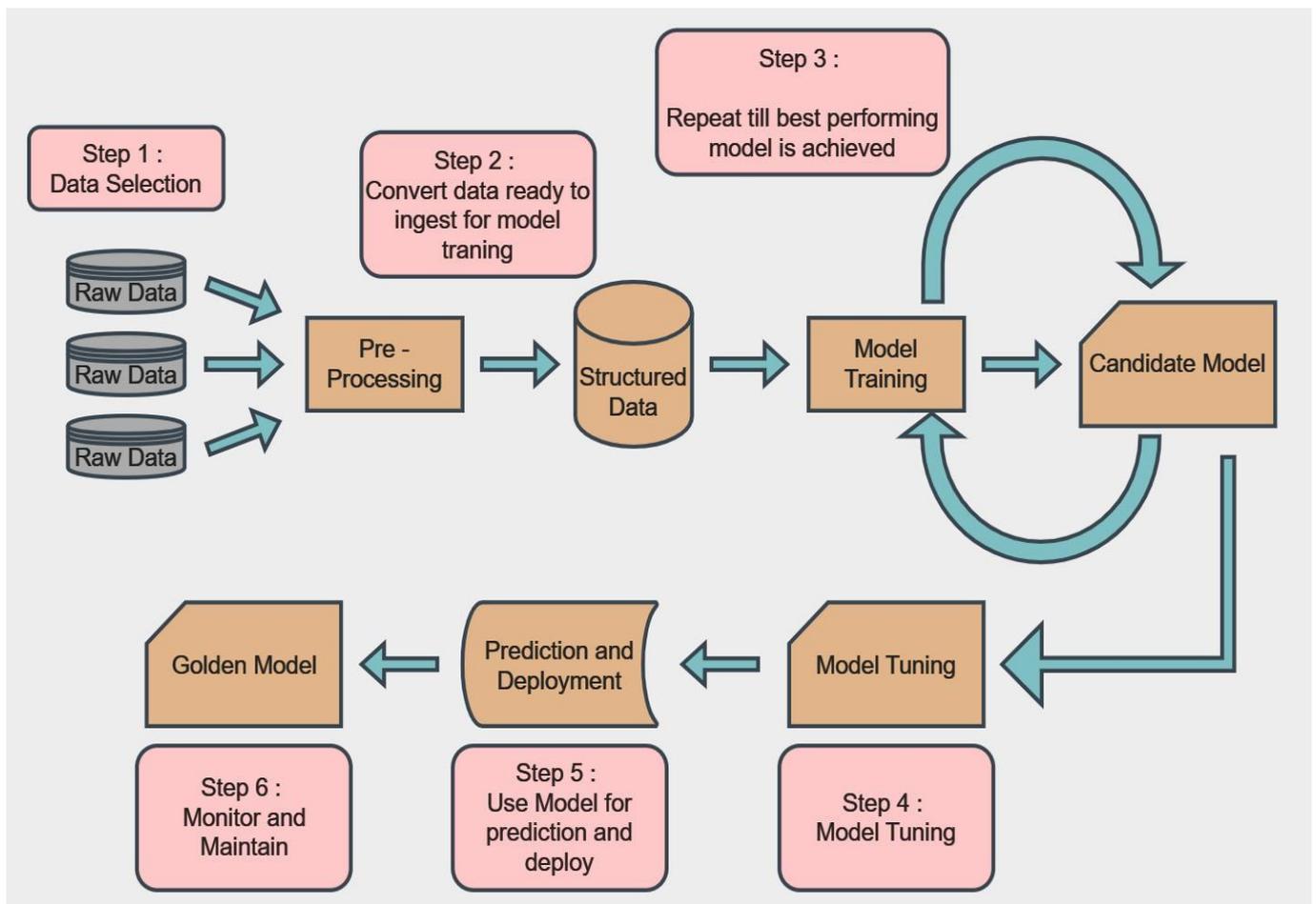

**Figure 1.** Key stages of ML implementation (our study).

This paper adopts the key stages presented in Figure 1, and each of these stages is discussed as follows.

Step 1: Data Collection. To effectively train ML models, a significant volume of data is essential, making it crucial to gather an ample amount of data for the project. It is essential to possess adequate domain knowledge to ensure that the collected data meet the project's requirements. The data may come in structured or unstructured forms, and they might contain missing, duplicate, or even incorrect values due to their diverse sources. Therefore, data preparation becomes a necessary step before applying ML algorithms.

Step 2: Data Preprocessing. During this stage, several data quality checks are performed to ensure the integrity of the data, including:

- Validity: Columns should contain relevant data types, such as integers, Booleans, or dates, with dates being in the correct format and within the required range. Mandatory fields in columns should not be empty. If there is no column with unique values to serve as an identifier, one can be created.
- Data Cleaning: This involves addressing missing values by either dropping observations or replacing the missing values with suitable alternatives. Duplicate data should be removed. Values may need to be transformed to formats suitable for ML algorithms.
- Denoising: The goal here is to remove errors and reduce variability in variables. Techniques such as binning, regression, and clustering can be used to achieve this.
- Outliers: Identifying and handling outliers is crucial. Outliers are values in columns that do not fall within the normal cluster or group. Outliers can have a significant impact on prediction outcomes (often negatively), so it is important to identify and appropriately address them.

Step 3: Model Training. The pre-processed data are structured and ready for the application of ML. The ML process involves iterating through different models and recording and evaluating their outputs, referred to as candidate models. A candidate model represents a prospective algorithm or architectural configuration under consideration for addressing a specific problem. ML practitioners often assess numerous candidate models to determine which one performs optimally for their particular task. This evaluation involves training and evaluating each model with relevant data and employing metrics like accuracy or loss functions. The final selection hinges on factors such as model accuracy, computational efficiency, interpretability, and resource constraints. In some cases, hyperparameter tuning may be applied to further enhance performance. The choice of a candidate model ultimately aims to strike a balance between the problem's requirements and the available computational resources and expertise.

Step 4: Model Tuning. Finding the optimal values for a model's configuration parameters (or hyperparameters) is the process of "model tuning" in ML. It entails deciding which hyperparameters to optimize, creating a space for their search, and using methods like grid search, random search, or Bayesian optimization to find the ideal combination. Choosing the best hyperparameters will help the model perform better on a particular task.

Step 5: Prediction and Deployment. Prediction and deployment are pivotal stages in the ML model lifecycle. Prediction entails utilizing a trained model to make informed forecasts or classifications on new unseen data. This could involve categorizing emails as spam or not, predicting stock prices, or diagnosing diseases based on medical images. Deployment, on the other hand, involves the integration of a trained model into a practical application or system. It requires careful considerations for factors like scalability, latency, and security. Deployed models could be used in recommendation systems, autonomous vehicles, fraud detection, and countless other real-world scenarios, playing a crucial role in leveraging ML for practical decision making and automation.

Step 6: Monitoring and Maintain. After the system has been deployed and final checks have been conducted to ensure optimal performance, the model is designated as the "Golden model". A golden model can refer to an established trusted version of a model that serves as a performance reference point. This reference model embodies the model's original known good state and behavior. ML models can drift over time due to changing data distributions or other factors, impacting their accuracy and reliability. By continually comparing the current model's performance to the golden model's expected results, deviations or performance degradation can be detected. When significant discrepancies arise, it signals the need for model retraining or maintenance to bring the model's performance back in line with the desired standards, ensuring ongoing effectiveness and reliability in real-world applications.

*3.7. Model Evaluation*

This section delves into the evaluation metrics of the ML models applied to predict house prices. The goal is to provide a thorough understanding of the metrics used for the model comparisons and the selection of the best-performing model. We shall thus discuss the evaluation metrics such as R-squared, adjusted R-squared, mean absolute error (MAE), mean square error (MSE), and root mean squared error (RMSE).

The R-squared value serves as an indicator of how effectively the model is fitted and the accuracy of its predictions on unseen data samples. To calculate the R-squared value, you can use the following equation (formula for R-squared value)

$$R^2 = 1 - \frac{SSR}{SSM} \tag{13}$$

where SSR is the squared sum of error of regression line and SSM is the squared sum error of mean line.

The adjusted R-squared metric, which also assesses the model's goodness of fit, once again, XGBoost shines with a strong adjusted R-squared value. The formula for adjusted R-squared is given in Equation (14):

$$\text{Adjusted } R^2 = 1 - \frac{n-1}{n-k-1} * 1 - R^2 \tag{14}$$

where n is the number of observations and k is the number of independent variables

The mean square error (MSE) is calculated by determining the squared difference between actual and predicted values. A lower MSE signifies a better-performing model. The equation below presents the formula to calculate the MSE of the model.

Furthermore, an extension of MSE is the root mean squared error (RMSE), which is essentially the square root of MSE. It serves as an additional metric to bolster our assessment. Once again, a lower RMSE suggests that the model's predictions align closely with the actual values.

$$MSE = \frac{1}{n} \Sigma \{y - \hat{y}\}^2 \tag{15}$$

Finally, the **mean absolute error (MAE)** is examined, a metric that quantifies the difference between predicted and observed values. It is computed as the sum of absolute errors divided by the sample size, as depicted in the equation below.

$$MAE(y, \hat{y}) = \frac{1}{n_{samples}} \sum_{i=0}^{n_{samples}-1} |yi - \hat{y}i^1| \tag{16}$$

A lower MAE score signifies a superior model.

*3.8. Hyperparameter Tuning*

To construct an optimal ML model, it is necessary to explore various possibilities. Hyperparameter tuning plays a crucial role in achieving the ideal model architecture and the optimal configuration of hyperparameters. In particular, for deep neural networks and tree-based ML models that involve numerous parameters, tuning them is essential for building an efficient ML model [41]. KerasTuner is a scalable and user-friendly framework designed for hyperparameter optimization, specifically tailored for TensorFlow, a leading ML platform. With KerasTuner, it becomes possible to finetune hyperparameters such as the number of convolution layers, the number of input neurons, and the optimal number of epochs [42]. On the other hand, GridSearchCV is a technique that utilizes hidden layer neurons to modify hyperparameters and find the best values for a specific model. When it comes to optimizing the parameters of the support vector engine and creating an SVM classification model, parameter tuning with GridSearchCV has demonstrated its effectiveness in providing reliable and efficient optimization suggestions [43]. Furthermore, RandomSearchCV is an alternative and is accessible using the sklearn module. The library provides a technique that optimizes by a cross-validated search over the parameter settings. In contrast to GridSearchCV, not all parameter values are tried out, but rather a fixed number of parameter settings is sampled from the specified distributions. The number of parameter settings that are tried is given by n_iter.

## 4. Results

From Table 3, it is clear that XGBoost stands out with the highest R-squared value, specifically at 0.93. Following that, XGBoost also boasts the highest cross-validation score (CV), sitting at an impressive 88.940. This score signifies the model's capacity to generalize well across the entire dataset, making XGBoost the most suitable regression technique for this specific dataset. Now, considering the model's accuracy, XGBoost not only leads the pack but also showcases exceptional performance without hyperparameter tuning. With the optimal parameters, its accuracy further improves. Moving on to the adjusted R-squared metric, which also assesses the model's goodness of fit, once again, XGBoost shines with a strong adjusted R-squared value. A lower MSE signifies a better-performing model. Notably, XGBoost exhibits the lowest MSE value in this scenario, standing at a remarkable 0.001. In addition, a lower RMSE suggests that the model's predictions align closely with the actual values.

**Table 3.** Comparative analysis of model performance.

| Regression Technique | Experimental Set Up | Model Score | R. Sq. | Adj. R.Sq. | MSE | RMSE | MAE | C V Score |
|---|---|---|---|---|---|---|---|---|
| Linear Regression | None | 0.935 | 0.910 | 0.899 | 0.017 | 0.130 | 0.075 | 87.530 |
| Multi-Layer Perceptron | random_state = 1 max_iter = 500 | 0.703 | 0.640 | 0.607 | 0.066 | 0.257 | 0.176 | 4.000 |
| Random forest Regression | n_estimators = 200 max_depth = 8 max_features = 9 | 0.880 | 0.860 | 0.850 | 0.025 | 0.159 | 0.112 | 82.320 |
| Support Vector Regression | None | 0.489 | 0.570 | 0.526 | 0.079 | 0.282 | 0.211 | 50.280 |
| **XGBoost** | **None** | **0.997** | **0.920** | **0.911** | **0.015** | **0.112** | **0.084** | **88.940** |

In an interesting exception, linear regression boasts the lowest MAE score, standing at 0.075, followed closely by XGBoost at 0.084. Despite linear regression's notable performance in terms of MAE in this instance, preference still leans towards XGBoost given its superior performance across all other metrics.

By examining the performance metrics, the significance of hyperparameter tuning becomes evident in Table 4 below. In this study, GridSearchCV was employed for parameter optimization. It is noteworthy that hyperparameter optimization yields consistent performance for the linear regression model, while it notably enhances the performance of both the random forest and XGBoost models. One standout example is the multi-layer perceptron, where the cross-validation score surges from 4 to an impressive 72, a substantial improvement. Furthermore, various other metrics exhibit enhancements due to the hyperparameter tuning implementation. However, there is an exception in the case of the support vector regression (SVR), where the CV score decreases from 50 to 0.87; most other metrics deteriorate as well, except for the model score, which sees a substantial increase. This could potentially be attributed to inadequate or incorrect parameter settings supplied to GridSearchCV or the SVR model's compatibility with the dataset. In summary, all models in the study demonstrate strong performances, except for SVR. Among them,

XGBoost emerges as the top-performing model, earning the title of the "Golden Model". Consequently, it is recommended for use in predicting house prices. Moreover, parameter optimization proves to be a valuable choice, significantly enhancing prediction accuracy. Therefore, the adoption of hyperparameter tuning is highly advisable for the ML regression models for optimality.

**Table 4.** An illustration of the impact of hyperparameter tuning on model performance.

| Regression Technique | Experimental Set Up | Model Score | R. Sq. | Adj. R. Sq. | MSE | RMSE | MAE | CV Score |
|---|---|---|---|---|---|---|---|---|
| Linear Regression | copy_x = True, fit_intercept = True, normalize = False | 0.935 | 0.910 | 0.899 | 0.017 | 0.130 | 0.075 | 87.530 |
| Multi-Layer Perceptron | random_state = 1, max_iter = 500 activation = tanh, hidden_layer_size = 50, solver = lbfgs, alpha = $5e^{-05}$ | 0.740 | 0.790 | 0.772 | 0.038 | 0.195 | 0.148 | 72.410 |
| Random Forest Regression | n_estimators = 300, max_depth = 10, max_features = 9 | 0.923 | 0.890 | 0.874 | 0.021 | 0.145 | 0.101 | 84.290 |
| Support Vector Regression | C = 5, verbose = 2, gamma = 0.1 | 0.946 | 0.020 | −0.081 | 0.181 | 0.425 | 0.332 | 0.870 |
| XGBoost | objective = "reg:squarederror", gamma = 0.001, max_depth = 10, min_child_weight = 50, sub = sample = 1 | **0.983** | **0.930** | **0.920** | **0.001** | **0.116** | **0.084** | **90.000** |

*Model Validation*

In the realm of ML, model validation is a crucial step involving the comparison of the test dataset with the trained model, as outlined by Alpaydin [44]. This validation process typically occurs after the model has undergone training. In the case of this study, which selected the XGBoost regression model, validation is conducted through the creation of scatter plots depicting actual values against predicted values as shown in Figure 2. Additionally, a histogram illustrating the differences between actual and predicted values is generated in Figure 3.

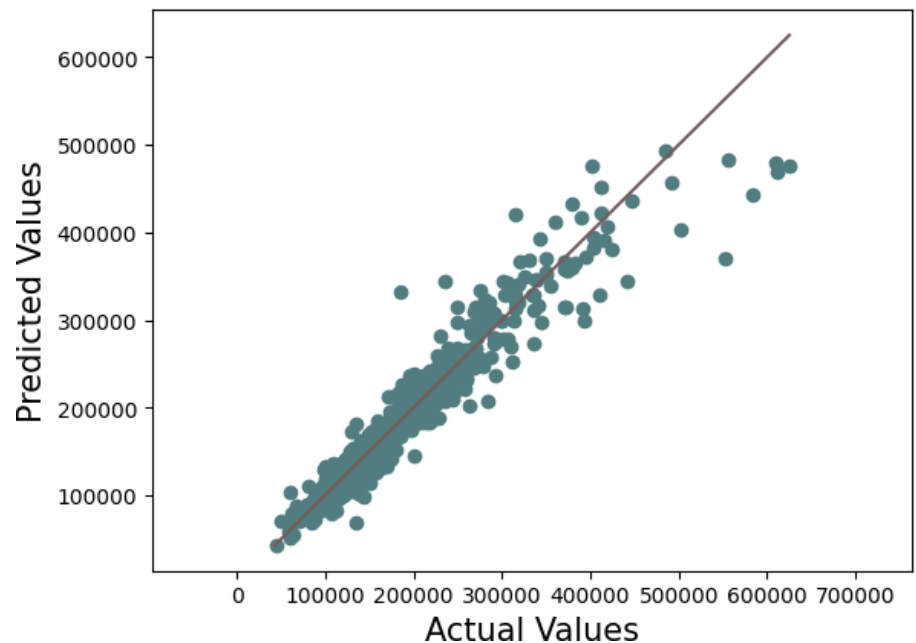

**Figure 2.** Scatter plot of actual vs. predicted values, XGBoost model.

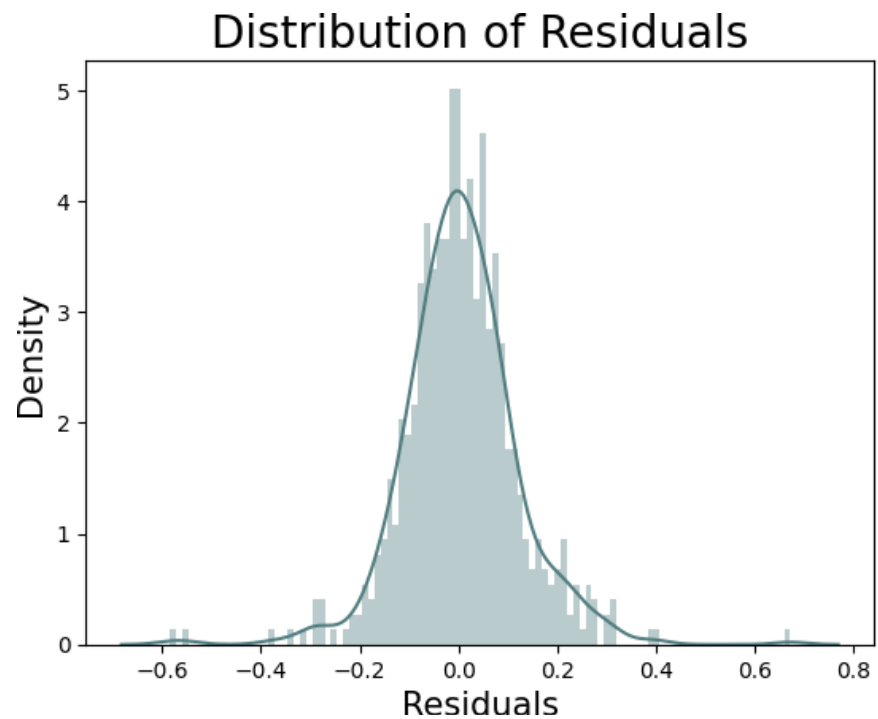

**Figure 3.** Histogram of actual predicted values, XGBoost model.

The scatter plot comparing actual and predicted values represents the most informative data visualization technique. Examining Figure 2 above, it becomes evident that the majority of data points closely align with the fitted line, with only a few outliers present. This observation strongly supports the notion that the model is indeed well-suited for making predictions on this dataset.

Next, we turn our attention to the histogram representing the residual plot for the XGBoost model. Upon examination, it becomes apparent that the distribution of values exhibits symmetry and conforms to a bell curve shape. This characteristic suggests that the distribution of residual values is normal in nature. Additionally, the narrowness of the curve indicates remarkably low deviation. Therefore, through the application of both visualization techniques, we can confidently affirm and bolster the validity of our chosen model. The concept of "feature importance" pertains to techniques that assign a score to each input feature within a model, reflecting the relative significance of each feature. A higher score signifies that a particular characteristic wields a more substantial influence on the model's ability to predict the target variable. In this research, feature selection is carried out using a random forest model. To determine the optimal features for constructing the random forest model, a feature selection methodology rooted in hypothesis integration is introduced. This approach has been empirically proven to yield excellent algorithm performance while reducing dimensionality [45]. In simpler terms, feature selection aids in identifying the key features—those variables that exert the most profound impact on the prediction model.

As depicted in Figure 4 below (generated by XGBoost's feature importance analysis), the feature with the most substantial influence on house prices is the overall quality of the house, closely followed by ground floor of the living area, garage cars, and total basement sq.ft. These factors play a significant role in determining house prices. This information holds practical value for both real estate firms and prospective buyers. Real estate construction companies can utilize these insights when designing homes, while house buyers can consider these variables to make informed purchasing decisions at equitable prices.

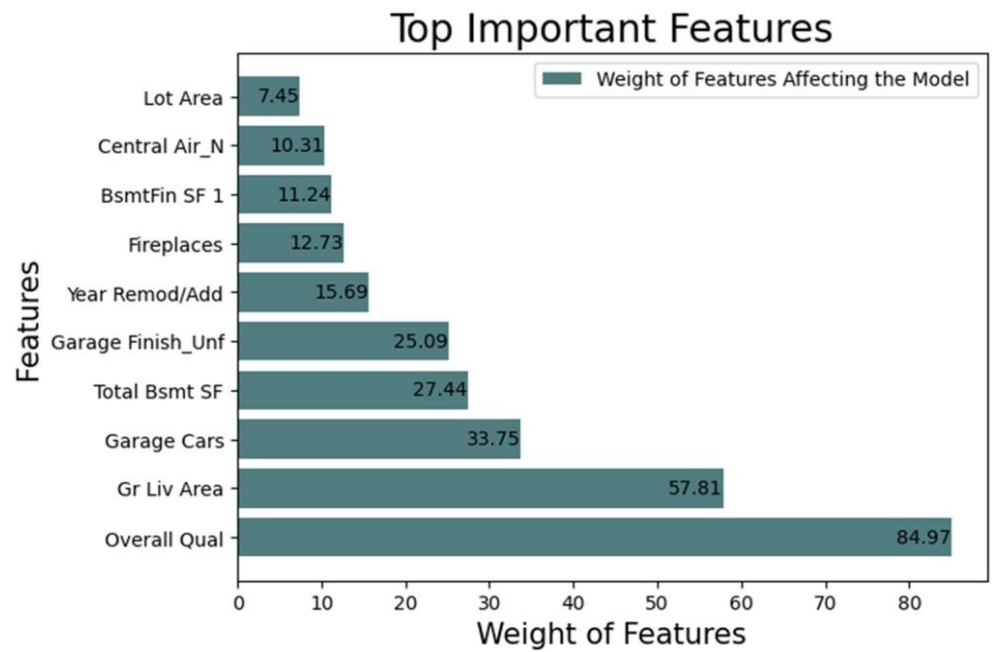

**Figure 4.** Feature importance with XGBoost.

## 5. Conclusions

In this study, we compared linear regression, multi-layer perceptron, random forest regressor, support vector regressor, and XGBoost. Afterwards, we demonstrated the use of GridSearchCV to achieve optimal solution by performing hyperparameter tuning of the ML algorithms. Our results showed that XGBoost is the optimal model for house price prediction, with a minimal MSE of 0.001. Furthermore, we utilized the power of ensemble trees (XGBoost) to identify the important features of our model. This helped us to understand the strength of the features in predicting house prices. In this case, we evidenced that the top important variables in predicting house prices are "*Overall Qual*" (overall quality of the house), "*Gr Liv Area*" (ground floor of the living area), "*Garage Cars*" (garage for the cars), and "*Total Bsmt SF*" (total basement sq.ft.). This study has effectively engineered a robust ML system for precise house price predictions and variable identification, providing practical value to real estate professionals, investors, and potential homebuyers. While recognizing certain limitations, primarily linked to data availability, it is expected that these challenges will gradually diminish as housing data become more digitally accessible. As a result, this research stands as an invaluable reference for future investigations in this domain, underlining XGBoost's consistent superiority and emphasizing the significance of key factors like overall quality and ground floor living area in informed decision making within the ever-evolving housing market. To conclude, the key findings from our study are as follows.

- We propose the use of the XGBoost algorithm as an interpretable, easy, simple, and more accurate model for house price prediction.
- Not all regression models perform equally well.
- Hyperparameter tuning consistently improves model performance, though not universally.
- GridSearchCV hyperparameter tuning proves beneficial for enhancing model performance.
- Identifying and emphasizing key influential features is crucial in house price prediction context.

*Limitations and Future Scope*

The primary focus for the future revolves around expanding the scope of this research while thoughtfully addressing its limitations. Several considerations for future enhancements include:

- Acquiring high-volume data to enhance accuracy.
- Extending data coverage to larger areas like states or entire countries.
- Utilizing data from verified and reliable sources.
- Employing multiple datasets to compare results and improve prediction accuracy and generalization.
- Exploring a wide range of methods once an optimal model is achieved.
- Leveraging the skills gained from this project and other sources.
- Studying additional research within the same field to enhance domain knowledge.
- Implementing spatial interpolation techniques for predicting house prices.
- Developing a standalone application for public access.

However, obtaining high-volume data poses challenges, including extensive computational time. In this study, with fewer than three thousand observations, hyperparameter tuning for one model took approximately 8 h. With millions of observations, this could be significantly longer. Solutions include limiting parameters and investing in high-performance workstations. Also to be considered is spatial interpolation, which factors in the proximity of facilities like parks, hospitals, and railway stations, all of which strongly influence house prices [22].

**Author Contributions:** Conceptualization, H.S., H.H. and B.O.; methodology, H.S., H.H. and B.O.; software, H.H.; validation, H.H.; formal analysis, H.H.; investigation, H.H.; resources H.S. and H.H.; data curation, H.S., H.H. and B.O.; writing—original draft preparation, H.S., H.H. and B.O.; writing—review and editing, H.S., H.H. and B.O.; visualization, H.S., H.H. and B.O.; supervision, H.S. and B.O.; project administration, H.S. and B.O. All authors have read and agreed to the published version of the manuscript.

**Funding:** This research received no external funding.

**Institutional Review Board Statement:** Not applicable.

**Informed Consent Statement:** Not applicable.

**Data Availability Statement:** The dataset and code used in this work are available via the links below. Dataset—https://www.kaggle.com/datasets/shashanknecrothapa/ames-housing-dataset (accessed on 9 November 2023). Code—https://github.com/hiteshharsora/housepriceprediction.git.

**Conflicts of Interest:** The authors declare no conflicts of interest.